\documentclass[10pt,twocolumn,letterpaper]{article}

\usepackage{iccv}
\usepackage{times}
\usepackage{epsfig}
\usepackage{graphicx}
\usepackage{amsmath}
\usepackage{amssymb}

\usepackage[pagebackref=true,breaklinks=true,letterpaper=true,colorlinks,bookmarks=false]{hyperref}

\iccvfinalcopy 

\def\iccvPaperID{754} 

\ificcvfinal\fi

\begin{document}

\title{Vision-Infused Deep Audio Inpainting}

\author{Hang Zhou$^{1}$ \quad Ziwei Liu$^{1}$ \quad Xudong Xu$^1$ \quad  Ping Luo$^{2}$ \quad Xiaogang Wang$^1$\\
    $^1$CUHK - SenseTime Joint Lab, The Chinese University of Hong Kong \\
    $^2$The University of Hong Kong\\
    {\tt\small \{zhouhang@link,xx018@ie,xgwang@ee\}.cuhk.edu.hk}\hspace{10pt}
    {\tt\small zwliu.hust@gmail.com}\hspace{10pt}
    {\tt\small pluo@cs.hku.hk}
}

\maketitle
\ificcvfinal\thispagestyle{empty}\fi

\begin{abstract}
    Multi-modality perception is essential to develop interactive intelligence. 
    In this work, we consider a new task of visual information-infused audio inpainting, \ie synthesizing missing audio segments that correspond to their accompanying videos. We identify two key aspects for a successful inpainter: (1) It is desirable to operate on spectrograms instead of raw audios. Recent advances in deep semantic image inpainting could be leveraged to go beyond the limitations of traditional audio inpainting. (2) To synthesize visually indicated audio, a visual-audio joint feature space needs to be learned with synchronization of audio and video. To facilitate a large-scale study, we collect a new multi-modality instrument-playing dataset called MUSIC-Extra-Solo (MUSICES) by enriching MUSIC dataset~\cite{zhao2018sound}. Extensive experiments demonstrate that our framework is capable of inpainting realistic and varying audio segments with or without visual contexts. More importantly, our synthesized audio segments are coherent with their video counterparts, showing the effectiveness of our proposed Vision-Infused Audio Inpainter (VIAI). Code, models, dataset and video results are available at \url{https://hangz-nju-cuhk.github.io/projects/AudioInpainting}.
\end{abstract}

\section{Introduction}

Audio-visual analysis provides valuable and complementary information that is crucial for comprehensively modeling sequential data.
Substantial progress has been achieved in recent years.
For example, it has been shown that the two modalities of audio and video can be transformed from one to the other~\cite{chen2017deep, hao2018cmcgan}, that is, from video to audio~\cite{chung2017you, chen2018lip} and from audio to video~\cite{ephrat2017vid2speech, zhou2019talking, zhou2017visual}. 

\begin{figure}[t!]
\centering
\includegraphics[width=1\linewidth]{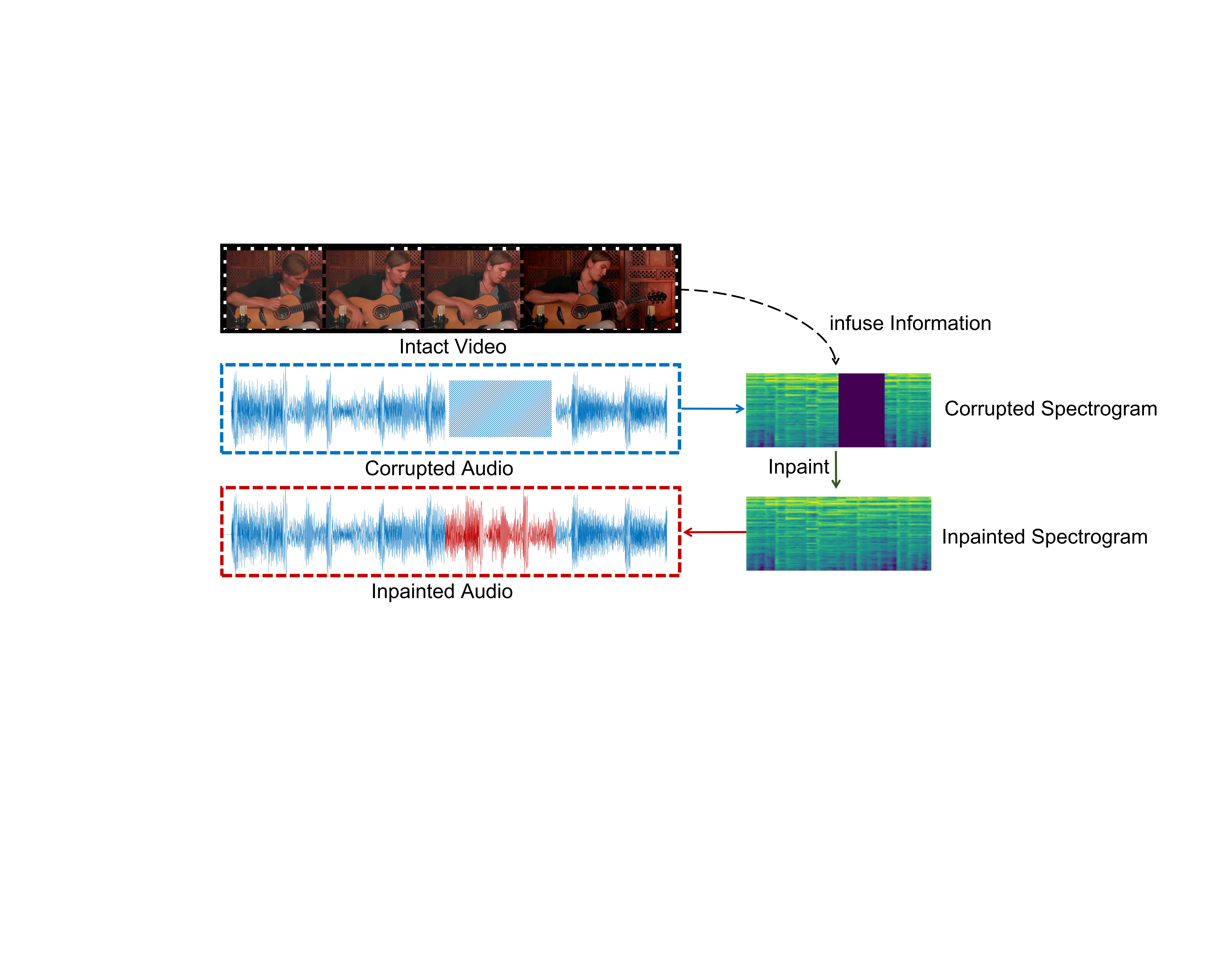}

\caption{Problem description. We study the problem of inpainting a clip of missing audio data, particularly with its corresponding video given. It is formulated into deep spectrogram inpainting, and video information is infused for generating coherent audio.}
\label{fig:intro}
\vspace{-10pt}
\end{figure}

This work focuses on a new task of audio inpainting, by using both video and audio as constraints. The inpainted audio segment is required to have the semantic concepts of the constraints, meaning that it has to be not only auditory reasonable but also visually coherent with the video. The setting of the problem is illustrated in Fig.~\ref{fig:intro}. 

In real life, audio signals often suffer from local distortions where the intervals are corrupted by impulsive noise and clicks. 
Even more, a clip of audio might be wiped out due to accident or transmission failure loss. 
To deal with such cases, a feasible operation is to fill the corrupted parts with newly generated samples, which can be referred to as audio inpainting~\cite{adler2012audio}.

While directly predicting a missing piece of audio is difficult, concrete information about audio signals could be provided by intact visual information accompanying the audio data.
The visual cue can be regarded as both a constraint and self-supervision to guide audio generation.  
%
In this paper, we present a vision-infused method that can deal with both audio-only and audio-visual associated inpainting.

%

Audio inpainting is significantly challenging as a consequence of audio's property of high sampling rate and long-range dependency. 
Traditional methods normally exploit the sparse representation of audio~\cite{adler2012audio, bahat2015self, chantas2018sparse, perraudin2018inpainting, toumi2018sparse}, and seek to find similar signal structures. 
However, similar structures do not always exist in the given inputs, especially when the inputs are short.
Moreover, most of the previous work cannot handle missing lengths longer than 0.25 seconds~\cite{bahat2015self}. 
And neither are these methods able to associate with given videos.

Another idea is to apply recent advances in audio generative tasks by using deep learning.
A recent work that closely related to ours is~\cite{zhou2017visual}, which uses videos as conditions to directly generate audio signals. However, previous methods have not explored the smoothness constraint on both sides of the to-be-inpainted audio.

To tackle these problems, our key insight is that \emph{we can effectively exploit the context information in audio by viewing the compact audio representation of spectrogram as a continuous signal}. Inspired by recent deep models of image inpainting~\cite{pathak2016context, iizuka2017globally, yu2018generative}, we formulate the problem in the same way, regarding spectrogram as a special kind of ``image", by treating time and frequency as height and width. Researchers have shown that spectrogram can be effectively processed by convolutional neural networks (CNNs)~\cite{chung2017lip, zhao2018sound}. 
%
%
%
 We believe a convolutional encoder-decoder network is able to recover high-level timbre and low-level frequency of the missing audio parts. This requires the spectrogram to contain enough yet simple information. With this motivation, we use the representation of Mel-spectrogram and design a spectrogram inpainting pipeline with generative adversarial networks (GAN)~\cite{goodfellow2014generative}.

We then incorporate visual information into this pipeline. 
We propose the core of extracting desired information is to find a joint feature space where audio and video are synchronized so that the shared rhythm information could be provided to the network. Finally, a WaveNet~\cite{van2016wavenet} decoder with mixture logistic loss is trained to recover high-quality audio from the spectrogram for the target source (instruments for music). The WaveNet decoder also benefits us at utilizing previous clean data. Since our spectrogram inpainting pipeline is inspired by the computer vision community, and the model itself is designed to be able to extend to audio-visual version, the proposed audio inpainting system is, in principle, infused by visual signal. Therefore we formally term our framework as Vision-Infused Audio Inpainter (VIAI). 


Our \textbf{contributions} are summarized as follows.
(1) We propose a novel framework for audio inpainting inspired by image inpainting to perform on spectrograms. An inpainted spectrogram is then converted into coherent audio with a WaveNet decoder.
(2) We incorporate visual cues into this framework and, to the best of our knowledge, design the first system targeting video-associated audio inpainting. 
(3) Along with our model, we also introduce novel training strategies for effective learning. Extensive experiments show that our framework can successfully handle missing music clips at lengths around 0.8 seconds with only 4 seconds inputs. Such lengths cannot be handled by most of the existing audio inpainting methods.
(4) We extend the original MUSIC dataset~\cite{zhao2018sound} to a richer version, named MUSICES, to benefit the entire audio-visual research community.


\section{Related Work}

\noindent
\textbf{Audio Inpainting.}
Previous research mainly resolves audio inpainting from a signal processing point of view. Sparse approximation in the time-frequency domain has been explored in~\cite{adler2012audio, siedenburg2013audio}, but silence will be introduced when gap exceeds 50ms. Self-similarity has been employed to inpaint gaps up to 0.25 seconds using time-evolving features~\cite{bahat2015self}. 
Recently, using similarity graphs, \cite{perraudin2018inpainting} proposes to inpaint long music segments, but it cannot handle segments shorter than 3 seconds. More importantly, similar frames do not exist for certain in the given intact input areas. This kind of method would fail when such cases are presented. Only very recently, some contemporary works exploit CNNs for audio inpainting~\cite{marafioti2019audio}.

\noindent
\textbf{Audio Synthesis.}
By applying deep learning, generative models such as SampleRNN~\cite{mehri2016samplernn}, WaveNet~\cite{van2016wavenet} and their variants~\cite{engel2017neural, oord2017parallel} have successfully generated high fidelity raw audio samples. One of the most important developments is to use them as decoders for conditional audio generation tasks such as Text-to-Speech Synthesis (TTS). For example, acoustic features designed by domain expertise have been used as inputs for audio synthesis based on SampleRNN~\cite{ai2018samplernn} and WaveNet~\cite{arik2017deep, gibiansky2017deep}. Latter in Deep Voice 3~\cite{ping2017deep} and Tacotron 2~\cite{shen2018natural}, Mel-spectrogram has been successfully used to train WaveNets. Inspired by their works, we adopt a similar structure to generate raw audios in the proposed task.

\noindent
\textbf{Audio-Visual Joint Analysis.}
%
Recent years witness the rapid growth in audio-visual joint learning tasks such as audio-visual speech recognition~\cite{chung2016lip, chung2017lip}, learning audio-visual correspondence~\cite{arandjelovic2017look, arandjelovic2018objects, aytar2016soundnet}, localization~\cite{zhao2018sound, senocak2018learning}, synchronization~\cite{Chung16a, owens2018audio, korbar2018co}, audio to visual generation~\cite{chung2017you, zhou2019talking, hao2018cmcgan}, visual to audio generation~\cite{ephrat2017vid2speech, zhou2017visual, owens2016visually},  visually aided source separation~\cite{owens2018audio, ephrat2018looking, gao2018learning,zhao2019sound,xu2019recursive}, and spatial audio generation~\cite{gao2018visualsound, morgadoNIPS18}. 

Among them, works that map visual to sound \ie source separation and sound generation are more related to ours. Source separation works perform more often on spectrograms. Zhao \textit{et al.}~\cite{zhao2018sound} use Short-Time-Fourier-Transforms (STFT) to realize source localization and separation. Similarly, Ephrat \textit{et al.}~\cite{ephrat2018looking} use talking face videos to form masks on STFTs of speech signals to achieve speech separation. Owens and Efros~\cite{owens2018audio}, on the other hand, concatenate visual features in the bottleneck of a spectrogram U-net. 
Unlike source separation that all audio information to be recovered has already existed, generating new audio would be much more difficult. In~\cite{owens2016visually}, hitting sound is predicted specifically. And~\cite{zhou2017visual} directly generated sound for in the wild videos using SampleRNN. But our work has a different setting from all of them.

\begin{figure*}[t!]
\centering
\includegraphics[width=1\linewidth]{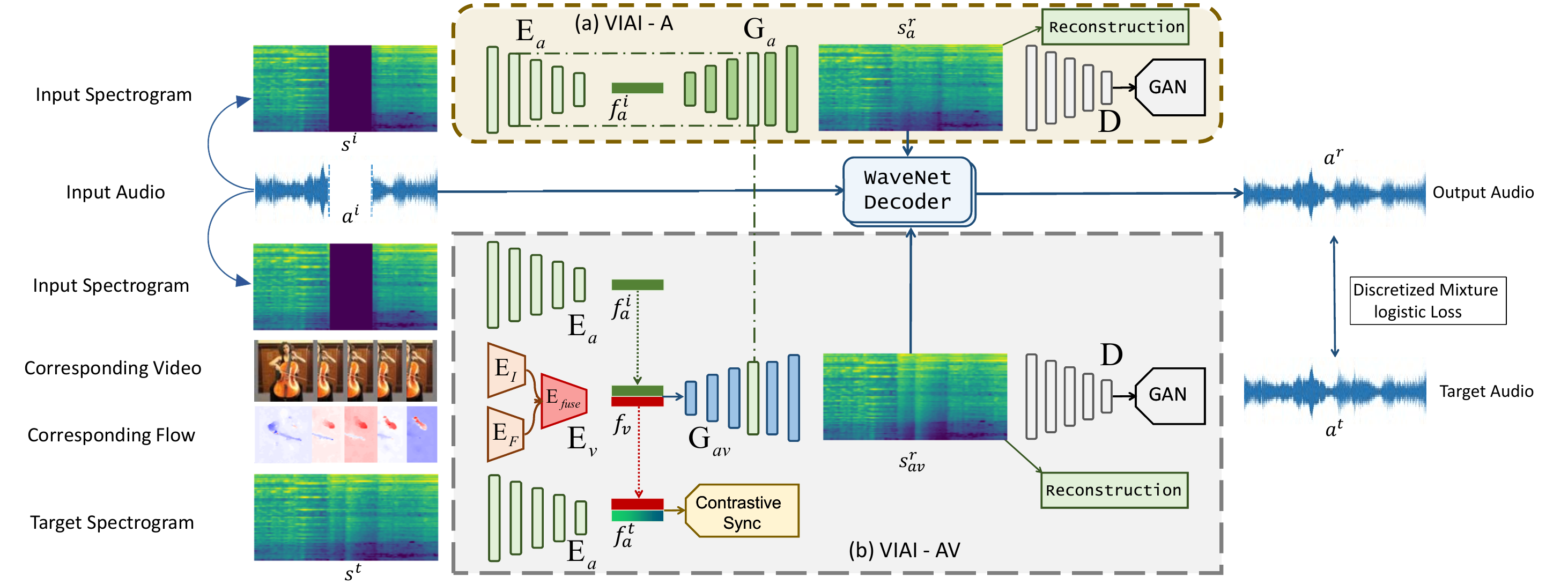}

\caption{The whole Vision-Infused Audio Inpainter system pipeline. In the above bracket (a) is the VIAI-A inpainting schedule. First the input corrupted audio is processed into Mel-spectrogram with a missing piece. An encoder-decoder pair \{$\text{E}_a$, $\text{G}_a$\} with one skip connection at the second layer restores the spectrogram to a complete one $s^r_a$. Below (b) is the VIAI-AV pipeline. Bottleneck features $f_{a}^t$, $f_v$ are extracted from audio and visual encoder $\text{E}_a$ and $\text{E}_v$. They are trained to be synchronized with each other. At the same time, concatenating $f_v$ with the distorted audio feature $f_a^i$ from $\text{E}_a$, the decoder $\text{G}_{av}$ reconstruct the spectrogram $s^r_{av}$ base on both information. The reconstruction output results $s^r_{a}$ and $s^r_{av}$ are constrained with reconstruction loss and GAN loss with the target $s^t$. Finally the results are sent into the pretrained WaveNet decoder to generate raw audio.}
\label{fig:pipeline}
\end{figure*}

\noindent
\textbf{Image Inpainting.}
Image inpainting~\cite{pathak2016context,yeh2017semantic} is a well-studied topic in computer vision and graphics. 
Deep learning methods have been successfully applied to this field with GANs. Context Encoders~\cite{pathak2016context} firstly trains deep encoder-decoder networks for inpainting with large holes. \cite{iizuka2017globally} extends it with global and local discriminators as adversarial losses. Recently, researchers dig into the combination of deep learning methods and exemplar-based approaches~\cite{yan2018shift, wang2018image}. Same practice could be applied in our framework, but for simplification of our proposed method, we just borrow the encoder-decoder baseline.

\section{Our Approach}

We introduce our Vision-Infused Audio Inpainter (VIAI) in this section. VIAI consists of two parts, a pure audio module ``VIAI-Audio (VIAI-A)'', and an audio-visual joint inpainting module ``VIAI-Audio-Visual (VIAI-AV)''. They all share a modified WaveNet decoder.
Fig~\ref{fig:pipeline} depicts the entire pipelines. The main idea is to turn audio inpainting into spectrogram inpainting in an image inpainting style. We first borrow undistorted audio around the missing part to form an input audio segment $a^i$. Then it is transformed into its Mel-spectrogram representation $s^i$ given the missing data length and position. Our goal is to reconstruct a spectrogram $s^r$, which is as similar to the target one $s^t$ as possible.

\subsection{Audio Inpainting as Spectrogram Inpainting}
\label{3.1}
\noindent\textbf{Pipeline.} The yellow bracket at the top of Fig~\ref{fig:pipeline} (a) shows the whole procedure of VIAI-A. We adopt an encoder-decoder architecture $\text{Net}_{a} = \{\text{E}_a; \text{G}_a\}$ with one skip connection. The bottleneck feature $f^i_a$ is a 1-d feature map (size of $1 \times \text{time} \times \text{channel}$), which gives the network the ability to deal with different input lengths. The output of the network is the reconstructed spectrogram $s^r_a = \text{Net}_a(s^i)$.

\noindent\textbf{Reconstruction.} While the skip connection benefits the network to directly take advantage of low-level information of the clean spectrogram by simple up-sampling operations, we design a weight adjusting training scheme to construct the missing part rely on high-level information from the bottleneck. Let $s^t_{\{m\}}$ be the target of the originally missing spectrogram parts and $s^r_{a\{m\}}$ be the predicted corresponding parts where $m$ denotes ``missing''. When applying the reconstruction $L_1$ loss, the weights between the originally clean and missing areas on the prediction and the target varies according to training time. The $L_1$ reconstruction loss can be written as:
\begin{align}
    \label{eq1}
{\mathcal {L}}^a_{re} = \eta_1 (t)\|s^t - s^r_a\|_1 + \|s^t_{\{m\}} - s^r_{a\{m\}}\|_1, 
\end{align}
where $\eta_1 (t)$ is a parameter which decays with the training steps, and set to a very small value after certain time. We find that if $\eta_1 (t)$ is fixed to 1, the network will learn mainly up-sampling. But if it is set to be very small at the first place, the network cannot restore the clean spectrogram clearly thus audio smoothness could be hurt.

Besides, a discriminator $\text{D}$ is trained with PatchGAN~\cite{pix2pix2017} objectives to maintain the local coherence and global similarity:
\begin{align}
    \label{eq2}
    {\mathcal {L}}^a_{\text{GAN}}(\text{Net}_a, D) =\ &  \mathbb{E}_{s^t}[\log \text{D}(s^t)]\ + \nonumber \\ 
    & \mathbb{E}_{s^i}[\log (1 - \text{D}(s^r_a)]
\end{align}
The total generation loss for VIAI-A is written as ${\mathcal {L}}^a_{Gen}$. $\beta$ is a hyper-parameter that leverages the two losses.
\begin{align}
    \label{eq3}
{\mathcal {L}}^a_{total} = {\mathcal {L}}^a_{Gen} = {\mathcal {L}}^a_{\text{GAN}} + \beta {\mathcal {L}}_{re}^a.
\end{align}

\subsection{Joint Visual-Audio Spectrogram Inpainting}
\label{3.2}
\noindent\textbf{Pipeline.} The pipeline for VIAI-AV is illustrated in the lower part (b) of Fig~\ref{fig:pipeline}. It evolves into a conditional inpainting problem by introducing the video encoder $\text{E}_v$ along with a synchronization module. The structure of audio encoder $\text{E}_a$ is kept unchanged. With the feature extracted by $\text{E}_v$ to be $f_v$, we aim to generate $s^r_{av} = \text{G}_{av}(\text{E}_a(s^i), f_v)$.

\noindent\textbf{Infusing Visual Cues.} The video corresponding to the target audio is provided. 
We believe motion information~\cite{liu2017video} is strongly associated with the change of audio melody, \ie, intense movements with rapid rhythms, so optical flows are extracted. Besides, \cite{zhou2017visual} shows that using both image and flow data can help improve direct audio generation results from videos. 
Each image and flow within this video are sent into encoder $\text{E}_v$, which contains ResNet encoders $\text{E}_I$, $\text{E}_F$ and down-sampling convolution layers $\text{E}_{fuse}$. Note that we control the down-sampling rate of $\text{E}_{fuse}$ to let $f_v$ match the size of $f_a$.

\noindent\textbf{Audio-Visual Synchronization.} Redundant information is contained in videos for audio reconstruction such as person appearances, the position of the instruments, and changing of background settings.
%
To capture the association between videos and audios, we propose to find a joint audio-visual space with synchronized rhythm information. 
In this joint space, visual feature $f_v$ is expected to be close to its corresponding intact target audio feature $f^t_a = \text{E}_a(s^t)$. 
%
We choose to use the contrastive loss as performed in~\cite{Chung16a, korbar2018co} that maps features into the same space.
The training objective is to minimize the distance between synchronized audio and video features and force the distance between unpaired data to be larger than a certain margin $\gamma$ :
\begin{align}
    \label{eq4}
    \mathcal{L}_{Sync} =& \sum_{n=1}^{N}{\|f^t_{a(n)} - f_{v(n)} \|_2^2 }~+ \nonumber \\ &\sum_{n\neq m}^{N, N}{\max(\gamma - \|f^t_{a(n)} - f_{v(m)} \|_2 , 0)^2},
\end{align}
where $N$ is the number of data in one batch, and $\gamma$ is set to 1. All the features are normalized first before implementation. The negative samples are drawn in a similar way as~\cite{korbar2018co}.

\noindent\textbf{Reconstruction with Probe Loss.} The video feature $f_v$ is then concatenated with the distorted bottleneck audio feature $f^i_a$ to form $f_{av}$, and sent to the new audio-visual decoder $\text{G}_{av}$ for spectrogram reconstruction $s^r_{av}$. The training objective is the same as section~\ref{3.1}, only substituting the subscript from $a$ to $av$ and get the generation loss: $\mathcal{L}^{av}_{Gen} = {\mathcal {L}}^{av}_{\text{GAN}} + \beta {\mathcal {L}}_{re}^{av}$.

In this video-associated scenario, crucial information about the missing piece is expected to be extracted from the condition feature $f_v$ by the decoder $\text{G}_{av}$. As $f^t_a$ is the compression of the clean spectrogram, information recovery from $f^t_a$ is easier and more obvious. So we reconstruct $s^r_{aa'} = \text{G}_{av}(\text{E}_a(s^i), f^t_a)$ using a similar $\mathcal{L}^{aa'}_{Gen}$ as a \emph{probe loss} to guide the learning of the networks. The idea is that while we restrict $f_v \approx f^t_a$ by applying the synchronization loss, we can suppose $\text{G}_{av}(\text{E}_a(s^i), f_v) \approx \text{G}_{av}(\text{E}_a(s^i), f^t_a)$. The process of this additional clean-audio-based inpainting module can be specifically named as VIAI-AA'. The success of generating terrific results with VIAI-AA' also proofs the ability of passing information from the bottleneck to the output. 


The overall objective of VIAI-AV can be written as:
\begin{align}
    \label{eq5}
{\mathcal {L}}^{av}_{total} = \eta_2(t){\mathcal {L}}^{aa'}_{Gen} + {\mathcal {L}}^{av}_{Gen} + {\mathcal {L}}_{Sync}.
\end{align}
$\eta_2(t)$ is a decay parameter that is similar with $\eta_1(t)$. 

\subsection{Spectrogram to Audio}

\noindent
\textbf{WaveNet Decoder.}
At the end of VIAI, a WaveNet decoder is attached for both the branches. Our choice of Mel-spectrogram is a way of data compression. With less information to recover for spectrogram inpainting, it is more complicate to transform it back into raw audio signals. So we utilize a modified version of the WaveNet architecture~\cite{van2016wavenet} to decode spectrogram into raw audio samples. WaveNet is an autoregressive model that is composed of dilated convolutions and non-linear activations. During training, it can take raw audio data as input and Mel-spectrogram as temporal conditions to predict the next-time-step audio in a teacher-forcing way. The Mel-spectrogram is first processed using up-sampling convolutions to match the sampling rate of raw audio data. During inference, WaveNet takes in one raw audio and upsampled spectrogram data at each time step, and generates the next time step's raw audio data. It models the conditional distribution between audio data and spectrogram $p({\bf{a}}|{\bf{s}})$:
\begin{align}
    \label{eq6}
p({\bf{a}}|{\bf{s}}) = \prod^{T}_{t = 1}p(a_{(t)}|a_{(1)},\cdots,a_{(t-1)}, s_{(t)})
\end{align}

We follow Parallel WaveNet~\cite{oord2017parallel} and Tacotran 2~\cite{shen2018natural} to use the discretized mixture logistic loss for training. One WaveNet model is pretrained for each class using clean audio samples and Mel-spectrograms in the dataset. A uniform WaveNet can also be trained in the same manner of multi-speaker TTS. 

\noindent
\textbf{Conditioning on Past Audio.}
In audio inpainting task, instead of simply modeling $p({\bf{a}}^r|{\bf{s}}^r)$, we take advantage of the WaveNet to rely the generation on both spectrogram and previous clean samples to model $p({\bf{a}}^r|{\bf{s}}^r, {\bf{a}}^i)$. Suppose the audio data is missing from time step $t_0$ to $T$, the distribution we model for the reconstructed audio $a^r$ given existing input audio $a^i$ and reconstructed spectrogram $s^r$ at time step $t$ ($t > t_0$) can be written as:
\begin{align}
    \label{eq7}
p({\bf{a}}^r|{\bf{s}}^r, {\bf{a}}^i) = \prod^{T}_{t = t_0}p(a^r_{(t)}&|a^i_{(1)},\cdots,a^i_{(t_0-1)}, \\ \nonumber
 & a^r_{(t_0)}, \cdots, a^r_{(t - 1)}, s^r_{(t)})
\end{align}
Finally, this customize WaveNet decoder can be integrated into our framework to constitute an end-to-end raw audio inference and training system.

\begin{figure}[t!]
\centering
\includegraphics[width=1\linewidth]{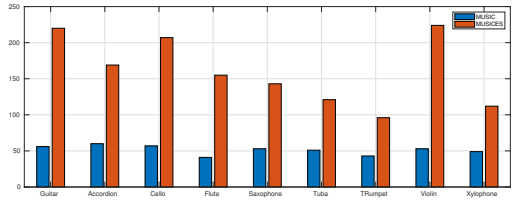}

\caption{Data statistic comparing with the original MUSIC dataset. The x-axis is the class name and the y-axis is the number of videos per class.}
\label{fig:stas}
\end{figure}

\section{MUSIC-Extra-Solo Dataset}
\noindent\textbf{Selection and Organization.} The strong association between audio and video can usually be found in videos of instrument playing. For example, the positions of hands and movements of the bow on strings can cast certain audio notes. But normally people cannot analysis the music notes according to only visual information. This provides our task with a suitable and challenging data option, so we turn to the recently proposed MUSIC dataset~\cite{zhao2018sound}. However, the released version has only around 50 videos for each class, which is not sufficient. Therefore, we extend the MUSIC dataset to approximately triple its original size on 9 of its major instruments. The additional videos are all solos, thus our extension is called the MUSIC-Extra-Solo (MUSICES) dataset. The statistics of the new dataset compared to the original one are summarized in Fig.~\ref{fig:stas}.

\noindent\textbf{Realistic Recorded Data.} Note that different from artificial music data generated with digital inference software such as MIDI, and videos recorded in a controlled lab environment, music data in MUSICES are mostly home camera recorded with certain background noise, which cast great difficulty for audio generation. The data are selected to be stable with good quality. In the original MUSIC dataset, important movements in certain videos could be invisible. This kind of video is kept out in our dataset. Different acoustic recording environments lead to domain differences~\cite{liu2019compound} of audios even in one single class, posing great challenge to our task.

\noindent\textbf{Detecting Video Shots.}  We also detect the shots changing within the dataset and provide the begin and end time of each shot. We observe that videos may contain black transition frames and clips that are silent before the player starts playing. So we split the videos according to our detected shots and abandon those non-auditory ones. Besides, the first 6 seconds of each video is cut out for data cleaning. Note that the train/test sets are divided first before cutting the videos in shots.

\noindent\textbf{Set-Splitting Protocol.} The train/test split is performed at the video level. Specifically, we split 10\% of the videos as a fixed testing set and randomly sampled 5\% as a held-out validation set. 

\begin{figure}[t!]
\centering
\includegraphics[width=1\linewidth]{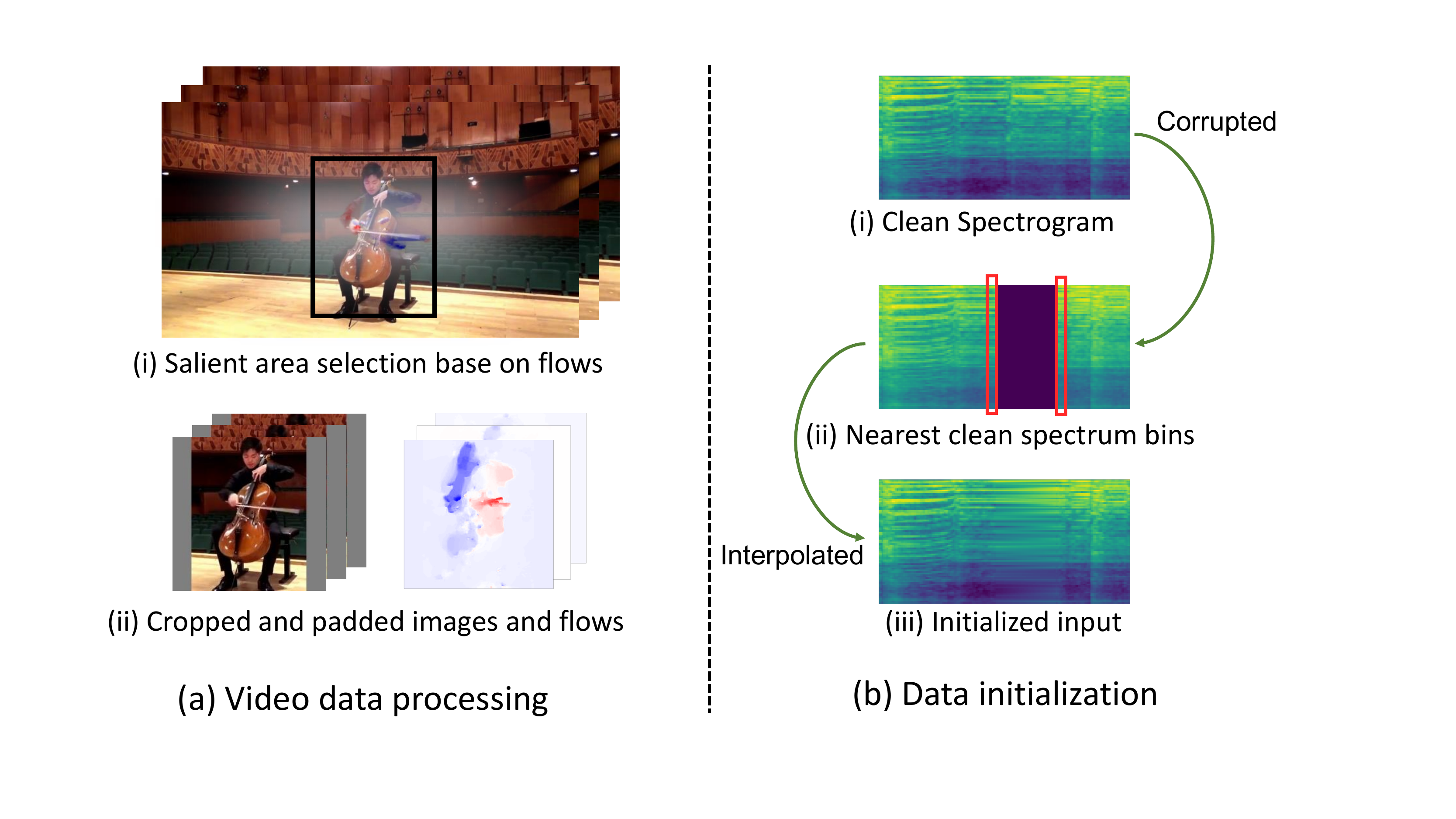}

\caption{Data pre-processing. (a) illustrate the procedure of flow-based salient area cropping for videos. (b) shows the results of spectrogram interpolating initialization}
\label{fig:dataset}
\end{figure}

\setlength{\tabcolsep}{4pt}
\begin{table*}[t] 
\small
\begin{center}

\label{table:psnr}
\begin{tabular}{cccccccc}
\hline
\tabcolsep=0.1cm

Score  $\setminus$ Approach & SampleRNN~\cite{mehri2016samplernn} & Visual2Sound~\cite{zhou2017visual}& bi-SampleRNN &bi-Visual2Sound  & VIAI-A & \textbf{VIAI-AV} & \textbf{VIAI-AA'} \\
\noalign{\smallskip}

\hline
PSNR   &9.1 & 10.2& 12.8 & 13.6&  22.2 &  \textbf{23.2}&  \textbf{26.6}\\
SSIM  &  0.33 &  0.35& 0.38 & 0.41 &  0.61 &  \textbf{0.64}&  \textbf{0.75}\\
 
\hline
SDR &  4.89 & 3.70& 4.20 & 4.72 &  6.54 &  \textbf{6.63}&  \textbf{6.89}\\
OPS &  51.1 & 51.3& 51.2 & 52.2 &  52.4 &  \textbf{56.3}&  \textbf{56.7}\\
\hline
\vspace{-15pt}
\end{tabular}
\end{center}
\caption{Quantitative results. The upper half are the evaluations of spectrograms and the lower half are the evaluation of audios. The maximum of OPS is 100. Larger values are better among these metrics. }
\end{table*}
\setlength{\tabcolsep}{1.4pt}

\section{Experiments}

\noindent\textbf{Data Processing.} Data processing is important in the realization of our approach, so we elaborate in this part. All audio samples are preprocessed to 16kHz sampling rate, then all raw audio amplitudes are normalized to between -1 and 1. Our Mel-spectrograms can be computed by firstly performing STFT using a frame length of 1280 points (corresponding to 80ms) and a hop size of 320 points (20ms). The STFT magnitude is transformed to Mel scale using an 80 channel Mel filterbank with a  frequency span from 125Hz to 7.6kHz, followed by log dynamic range compression. The spectrograms are normalized to between 0 and 1.

The spectrogram frame length and hop size are designed to map the 12.5 frame rate of its corresponding video. So temporally one video frame can be mapped to 4 spectrum bins. Optical flows are extracted by using TV-L-1 algorithm~\cite{zach2007duality} and bounded to be maximum 20 pixels. 
Salient areas inside a video are approximated by setting a threshold according to the average of all optical flow values over the video. 
Images and flows in one video are all cropped to this rectangular area with motion detected, and padded to be square. Fig.~\ref{fig:dataset} (a) depicts the procedure. Finally, the pixel values of images and flows are normalized to between -1 and 1. 

\noindent\textbf{Model Configurations.} The audio encoder $\text{E}_a$ consists of 5 stride-2 convolution layers with $3 \times 3$ kernels. 
The original 80 frequency bins are compressed to 1 by a final pooling layer.
Both the image and flow encoders $\text{E}_I$ and $\text{E}_F$ adopt the ResNet-18~\cite{he2016deep} architecture. One 256-length feature vector can be obtained from each image and flow. Then the features from one video clip are concatenated along the time axis. The following $\text{E}_{fuse}$ has two stride-2 1d convolutions. 

The decoder $\text{G}_a$ has 15 convolution layers with 6 bilinear upsample layers. The skip connection is at after the last upsample layer. Decoder $\text{G}_{av}$ different from $\text{G}_a$ only at the first convolution layer. It takes in twice the original feature length. As for the WaveNet decoder, we use 24 dilated convolution layers grouped into 3 dilation cycles instead of the original 30 layers for computational efficiency. One set of encoder-decoder and WaveNet model is trained for each class in the dataset.

\noindent\textbf{Experimental Settings.}
Throughout our experiments, we only consider missing lengths longer than traditional settings. When the missing length is short, differences between methods become difficult to be discriminated, and this problem is more challenging and realistic when the missing length is longer so this is the focus of our paper.

Our choice of training input data is 4s. The distortion is shorter than 1s but longer than 0.4s. The 4-second raw audio corresponds to an $80 \times 200$ size spectrogram and maps to 50 video frames. The bottle-neck feature map is extracted to be $256 \times 13$ with 13 to be the compressed time dimension.

\noindent\textbf{Implementation Details.}
During training, we manually crop a clip randomly within the clean spectrogram to create distortion. Different from image inpainting, the distorted part will be along the time axis (see Fig~\ref{fig:dataset} (b) for visualization of input spectrogram ). Based on the continuity of audio data, we initialize it to be the interpolation of the nearest clean spectrum bins as shown in Fig~\ref{fig:dataset} (b), instead of averaging ``pixel'' value like done in image inpainting. This interpolation-based initialization can directly lead to reasonable results under certain circumstances where the missing part is a stable music note but would fail in most cases.

Our implementation is based on PyTorch and trained on 4 Titan X GPUs. Networks are trained using Adam optimizer~\cite{kingma2014adam} with learning rate set to be 1e-4. The batch size is 64 when training VIAI-A and 16 for VIAI-AV. 
The decay parameter $\eta_1(t)$ and $\eta_2(t)$ are set to be $\max(0.1, 0.9^{step/1000})$. The synchronization loss $\mathcal{L}_{Sync}$ only updates video encoder $E_v$ as this stabilizes training.

\noindent\textbf{Competing Methods.}
We validate our spectrogram inpainting is superior to deep learning-based autoregressive audio generation methods with the listed baselines.  \textbf{SampleRNN}~\cite{mehri2016samplernn} has the ability to predict long-term audios with or without input conditions. We adopt it as an audio inpainting baseline. Then we reproduce \textbf{Visual2Sound}~\cite{zhou2017visual} as audio-visual baseline. Note that in original~\cite{zhou2017visual} paper, only ImageNet and action recognition pretrain network is used for feature extraction. For a fair comparison, we initialize their video extraction network to be our synchronizing pretrained ones. Also, we train an inverse SampleRNN model and fuse the outputs from both sides to create a \textbf{bi-directional SampleRNN} model. A similar \textbf{bi-Visual2Sound} model is also implemented. These approaches are compared with our method VIAI-A, VIAI-AV and a particular reference result of VIAI-AA', which is described in section~\ref{3.2} when reconstructing $s^r_{aa'}$. All experiments are conducted on the same set of data with the same pre-processing steps as described above. Note that we also reproduce state of the art traditional audio inpainting method which can handle the longest distortion~\cite{bahat2015self}, but it fails to generate any results on our setting. 

\begin{figure*}[t!]
\centering
\includegraphics[width=1\linewidth]{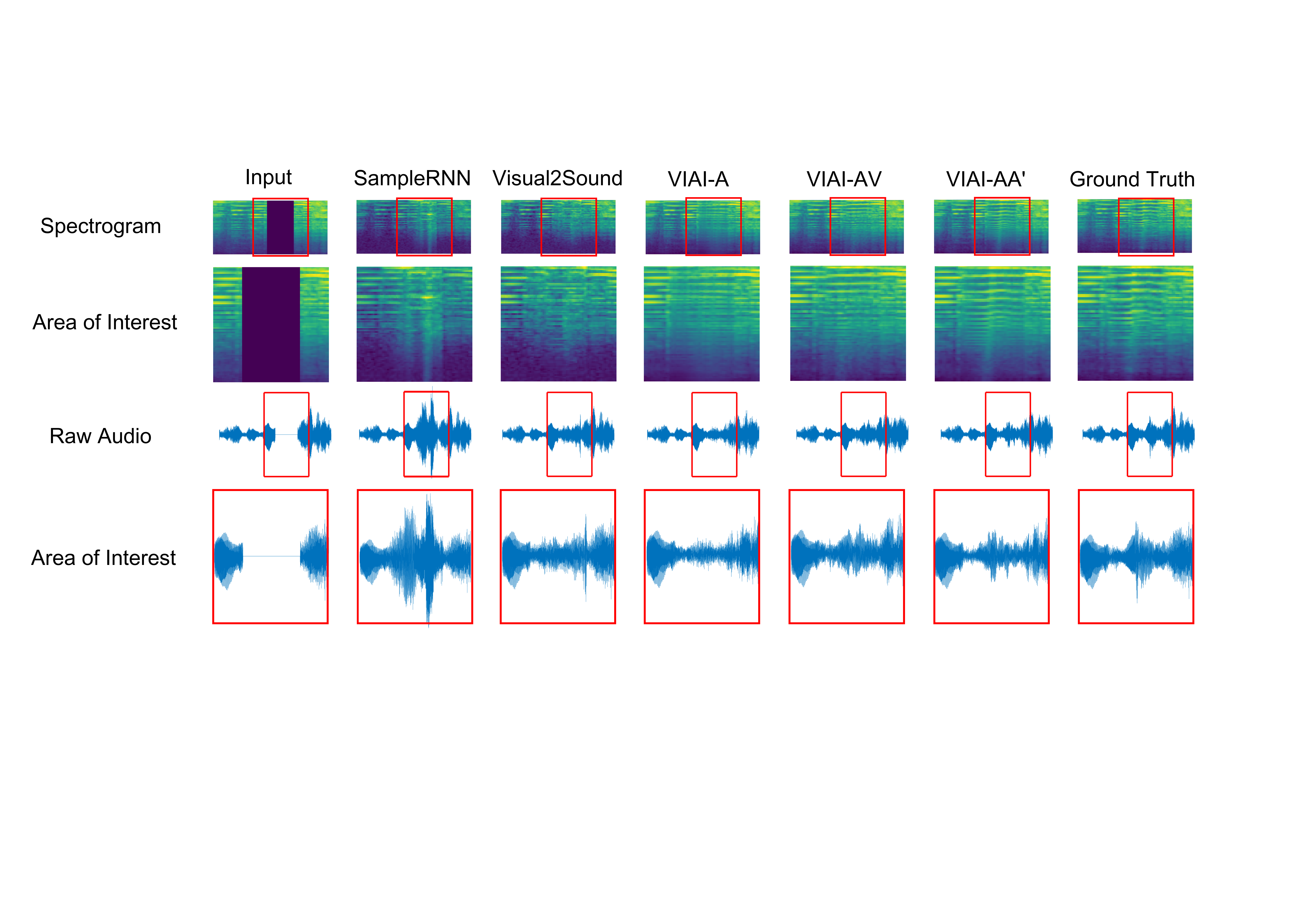}
\vspace{-10pt}
\caption{Qualitative results of a 0.8s distortion at an arbitrary position for different methods. The area of interest is the part in its corresponding red bracket above. Better viewed with zoom-in.}
\label{fig:qualitative}

\end{figure*}

\setlength{\tabcolsep}{4pt}
\begin{table*}[t] 
\begin{center}

\caption{Users' Mean Opinion Scores. Lager is higher, with the maximum value to be 5.}
\label{table:MOS}
\begin{tabular}{cccccc}

\hline

~~~~MOS on $\setminus$ Approach & SampleRNN~\cite{mehri2016samplernn} & Visual2Sound~\cite{zhou2017visual}  & ~~~~~~VIAI-A~~~~ & ~~~~\textbf{VIAI-AV} &~~~~ \textbf{VIAI-AA'}~~~~ \\
\noalign{\smallskip}
\hline

~~~~Audio Quality & 2.51 & 2.20 & ~~~~ 3.05 & ~~~~ \textbf{3.93}& ~~~~ \textbf{4.35}~~~~\\
~~~~Audio-Visual Coherence & 2.22 & 2.23 & ~~~~ 3.02 & ~~~~ \textbf{3.96}& ~~~~ \textbf{4.40}~~~~\\
~~~~Similarity with Target & 2.35 & 2.20& ~~~~ 2.97 & ~~~~ \textbf{4.01}& ~~~~ \textbf{4.46}~~~~\\
\hline
\vspace{-20pt}
\end{tabular}
\end{center}
\vspace{-10pt}
\end{table*}
\setlength{\tabcolsep}{1.4pt}

\setlength{\tabcolsep}{4pt}
\begin{table}[h] 
\caption{Users' MOS with bi-directional baselines.}
\vspace{-5pt}
\scriptsize
\tabcolsep=0.11cm
\begin{center}
\label{table:bi}
\begin{tabular}{ccccc}
\hline

MOS $\setminus$ Approach &  bi-SampleRNN  &  bi-Visual2Sound& VIAI-A  & \textbf{VIAI-AV} \\
\noalign{\smallskip}

\hline
Audio Quality & 2.89& 2.92&  3.12& \textbf{3.86}\\
Audio-Visual Coherence  & 2.76& 2.90& 3.05& \textbf{3.93}\\
Similarity with Target & 2.31& 2.65& 3.11& \textbf{3.96}\\
\hline
\vspace{-20pt}
\end{tabular}
\end{center}
\end{table}

\setlength{\tabcolsep}{1.4pt}

\subsection{Quantitative Evaluation}
\label{5.1}
Due to the limitation of paper length, we specifically show the results of \textbf{cello}, as a particular case study for quantitative evaluation. During the evaluation, the distorted length is fixed to be 0.8s for comparison. In the same way as training, the whole input audio length is selected to be 4 seconds. 
Only the missing area is considered under this setting, and 20 corrupted segments are sampled from each video in the test set for evaluation.

\noindent\textbf{Evaluation for Spectrograms.} We first evaluate the directly inpainted results of spectrograms by regarding them as images in the criterion of PSNR and SSIM~\cite{wang2004image} (larger is better). For our baselines~\cite{mehri2016samplernn} and \cite{zhou2017visual}, the audios are first generated then converted to Mel-spectrogram. 

\noindent\textbf{Evaluation for Audios.} We adopt audio evaluation protocols SDR and OPS from the audio-source separation community to evaluate the final inpainted raw audio results. SDR is the Signal to Distortion Ratio that directly comparing the data samples numerically. Defined in PEMO-Q auditory model~\cite{huber2006pemo}, 
OPS is the Overall Perceptual Score, which is also an objective assessment of audio quality proposed in~\cite{emiya2011subjective}. 

It can be observed from Fig.~\ref{table:psnr} that except for the model directly borrow intact audio information for inpainting (VIAI-AA'), results with video assistance surpass that of audio-only. And our VIAI system outperforms purely autoregressive models.

\subsection{Qualitative Evaluation}
\label{5.2}
We visualize a case in the form of spectrogram and raw audio at Fig~\ref{fig:qualitative}. The areas of interests are shown explicitly. The comparison shows that while autoregressive models fail to keep smoothness, our proposed VIAI-A generates visually reasonable and continuous results. Moreover, with the presence of visual information, our VIAI-AV model captures more details than VIAI-A. The result of VIAI-AA' reaches the best, which proves that information in the bottle-neck layer has indeed been used. For auditory results please refer to our video.

\noindent\textbf{User Study.}
Numerical numbers are hard to measure the true quality of audio signal, so we conduct a user study as a complimentary evaluation. The users are asked to evaluate the results with respect to the following three criteria; (1) Audio quality. The users mark how well the inpainting qualities are by listening to audios only. (2) Audio and Visual Coherence. To evaluate how well the inpainted audios are associated with the given videos. (3) The similarity to the ground truth. Compare the inpainted results with the ground truths and decide to what extent they are similar.

We utilize the widely used Mean Opinion Scores (MOS) rating protocol. There are overall 20 users taking part in the evaluation. The procedure for audio generation is the same as quantitative evaluation. We generate 50 different inpainted audio clips with all methods shown, and randomly assign 10 of them to one of the users. The users then give the ratings ranging from 1-5 with 5 to be the highest. Finally, all opinions are averaged.

The main results are listed in Table~\ref{table:MOS}. and results for bi-directional methods are conducted additionally, listed in Table~\ref{table:bi}. As illustrated, users prefer our VIAI system comparing to baselines by significant margins. Apparently, with video information infused, the system can inpaint audios that are coherent with their corresponding videos.

\setlength{\tabcolsep}{4pt}
\begin{table*}[t] 
\begin{center}

\begin{tabular}{ccccc}
\hline

~~~MOS on  $\setminus$ Approach ~~~& VIAI-AV' (no sync)~~~ & VIAI-AV (no prob)~~~ & VIAI-AV (no con)~~~ &\textbf{VIAI-AV}~~~ \\
\noalign{\smallskip}

\hline
~~~Audio Quality  ~~~& 2.90 & 3.59 & 3.00 &  ~\textbf{3.93}~~~ \\
~~~Audio-Viusal Coherence  ~~~& 2.95 & 3.65 & 3.17& \textbf{3.96}~~~\\
~~~Similarity with Target ~~&3.00 & 3.56 & 3.49 & \textbf{4.01}~~~\\
\hline
\vspace{-20pt}
\end{tabular}
\end{center}
\caption{Ablation study with Mean Opinion Scores.}
\label{table:ablation1}
\end{table*}

\setlength{\tabcolsep}{4pt}
\begin{table*}[h] 
\tabcolsep=0.1cm
\begin{center}

\begin{tabular}{lcccccccccccccccc}
\hline
Class &Violin & Accordion & Guitar& Flute& Xylophone&Trumpet&Saxophone &Tuba &Average\\

\hline
VIAI-A& 21.1$\mid$0.64 & 22.2$\mid$0.59 &21.3$\mid$0.58 & 22.4$\mid$0.60 & 20.2$\mid$0.56& ~~20.2$\mid$0.62& 21.5$\mid$0.59 & 20.0$\mid$0.57 & 21.2$\mid$0.60 \\
VIAI-AV& 22.4$\mid$0.66 & 23.6$\mid$0.61 & 21.9$\mid$0.61& 23.5$\mid$0.63 & 21.1$\mid$0.58& ~~21.0$\mid$0.64&22.5$\mid$0.60 &21.2$\mid$0.57 & 22.2$\mid$0.62\\
\hline
\vspace{-22pt}
\end{tabular}
\end{center}
\caption{PSNR$\mid$SSIM results on all classes.}
\label{table:results}
\vspace{-5pt}
\end{table*}
\setlength{\tabcolsep}{1.4pt}

\subsection{Ablation Study}

\noindent\textbf{Audio-Visual Synchronizing.} We propose that the audio-visual synchronizing part is the core of extracting desired visual information into the bottle-neck feature. Theoretically, the network will directly take the short-cut of the original VIAI-A path to inpaint base on spectrograms. We believe to use solely the reconstruction loss on VIAI-AV will render results similar to VIAI-A. The network trained without it is denoted by VIAI-AV'.

\noindent\textbf{Probe Loss of VIAI-AA'.} Then we investigate the help of the probe loss term $\mathcal{L}^{aa'}_{Gen}$.
Besides the already shown results in Section~\ref{5.1} and~\ref{5.2}, which demonstrate that latent information can be extracted from the bottle-neck, we further explore the influence of the existence of the loss term. The model is called VIAI-AV (no prob).

\noindent\textbf{Weight Adjusting for Reconstruction} To validate the effectiveness of the weight adjusting term $\eta_1(t)$ and interpolation initialization, we train extra experiments on AIVI-A by setting the coefficients of the loss term to be $\eta_1(0)$ and $\eta_1(+\infty)$. The experiment with the traditional fix value initialization is also performed as VIAI-A (old ini).

\noindent\textbf{WaveNet Conditioning.} Lastly, we use WaveNet to condition the generation on past results to further ensure smoothness. The training outcome without the conditioning term is addressed as VIAI-V (no con). 


\noindent\textbf{Ablation Results.} The results regarding the metric of PSNR and SSIM are shown in Table~\ref{table:ablation1}. Note that VIAI-AV (no con) shares the same inpainted spectrogram as VIAI-AV. 
We only perform subjective studies on these extra modified VIAI-AV methods at Table~\ref{table:ablation2}. As depicted in the tables, our final setting reaches optimal results regarding all kinds of criteria.

\setlength{\tabcolsep}{4pt}
\begin{table}[t] 
\begin{center}

\begin{tabular}{lcc}
\hline

~~~Approach  $\setminus$ Score & PSNR & SSIM~~~ \\
\noalign{\smallskip}

\hline
~~~VIAI-A~$\eta_1(0)$~~~ &  ~~~21.8~~~ & ~~~0.60~~~ \\
~~~VIAI-A~$\eta_1(+\infty)$~~~  &  ~~~21.6~~~ &~~~0.59~~~   \\
~~~VIAI-A (old ini)~~~ &  ~~~21.5~~~ & ~~~0.58~~~ \\
~~~\textbf{VIAI-A}~~~ & ~~~\textbf{22.2}~~~ & ~~~\textbf{0.61}~~~ \\
~~~VIAI-AV' (no sync)~~~ & ~~~21.8~~~ & ~~~0.62~~~ \\
~~~VIAI-AV (no prob)~~~ & ~~~22.5~~~ & ~~~0.63~~~ \\
~~~\textbf{VIAI-AV}~~~ & ~~~\textbf{23.2}~~~ & ~~~\textbf{0.64}~~~  \\
\hline
\end{tabular}
\caption{Ablation study with PSNR and SSIM metrics.}

\label{table:ablation2}
\end{center}

\vspace{-10pt}
\end{table}
\setlength{\tabcolsep}{1.4pt}

\subsection{Further Analysis}

\noindent\textbf{Analysis for Baselines.} The baseline methods are designed to generate continuous and reasonable results directly or following a probe input segment.
However, the task of inpainting requires the generated parts to be coherent with both sides of the existing audio parts. 
Particularly, Visual2Sound~\cite{zhou2017visual} fails to capture fine-grained visual information when applied to instrument playing data during our re-implementation. 

\noindent\textbf{Results on All Classes.} We conduct inpainting experiments on all 9 classes of our collected MUSICES dataset. The PSNR$\mid$SSIM results of the rest classes are shown in Table~\ref{table:results}.

\noindent\textbf{Failure Cases.} Failure could happen when the ground truth is already contaminated by noise, or the changing of music notes is too severe, which can be improved in the future.






\section{Conclusion}
In this paper, we have studied a new task and proposed an effective system called Vision-Infused Audio Inpainter (VIAI), which is capable of inpainting realistic and varying audio segments to fill in the corrupted audio. Our model integrates the intact corresponding video information into our framework to create inpainting results, which are coherent with the videos. Specifically, we formulate the problem of audio inpainting in the form of deep spectrogram semantic inpainting, and leverage the audio-visual synchronizing supervision to create a joint space for reconstruction. The novel usage of WaveNet decoder that conditions on both previous data and the reconstructed spectrogram enables the generation of high-quality raw audio data. Compared to prior methods, our approach can handle extreme inpainting settings that could not be processed by existing works, and it achieves audio-visual coherence audio-inpainting for the first time. Furthermore, an enhanced multi-modality dataset named MUSICES is contributed to the community for future audio-visual research. \\

\noindent\textbf{Acknowledgements.}
We thank Yu Liu and Yu Xiong for their helpful assistance. This work is supported in part by SenseTime Group Limited, and in part by the General Research Fund through the Research Grants Council of Hong Kong under Grants CUHK14202217, CUHK14203118, CUHK14205615, CUHK14207814, CUHK14213616.

{\small
\bibliographystyle{ieee_fullname}
\bibliography{egbib}
}

\end{document}